\documentclass[sigconf]{acmart}
\AtBeginDocument{%
  \providecommand\BibTeX{{%
    \normalfont B\kern-0.5em{\scshape i\kern-0.25em b}\kern-0.8em\TeX}}}

\copyrightyear{2022}
\acmYear{2022}
\setcopyright{acmlicensed}
\acmConference[MM '22]{Proceedings of the 30th ACM International Conference on Multimedia}{October 10--14, 2022}{Lisboa, Portugal}
\acmBooktitle{Proceedings of the 30th ACM International Conference on Multimedia (MM '22), October 10--14, 2022, Lisboa, Portugal}
\acmPrice{15.00}
\acmDOI{10.1145/3503161.3547804}
\acmISBN{978-1-4503-9203-7/22/10}

\usepackage{booktabs} 
\usepackage{tabularx}
\usepackage{multirow}
\usepackage{bm}
\begin{document}

\title{Cross-Modality High-Frequency Transformer for\\ MR Image Super-Resolution}

\author{Chaowei Fang}

\affiliation{%
  \institution{Xidian University}
  \country{}
}
\email{chaoweifang@outlook.com}

\author{Dingwen Zhang$^\ast$}
\affiliation{%
  \institution{Northwestern Polytechnical University}
  \country{}
  }
\email{zhangdingwen2006yyy@gmail.com}
\thanks{$^\ast$ Corresponding author}

\author{Liang Wang}
\affiliation{%
 \institution{Xidian University}
 \country{}
 }
\email{2451207324@qq.com}

\author{Yulun Zhang}
\affiliation{%
  \institution{Computer Vision Lab, ETH Zürich}
  \country{}
  }
\email{yulun100@gmail.com}

\author{Lechao Cheng}
\affiliation{%
  \institution{Zhejiang Lab}
  \country{}
  }
\email{chenglc@zhejianglab.com}

\author{Junwei Han}
\affiliation{%
  \institution{Northwestern Polytechnical University}
  \country{}
  }
\email{jhan@nwpu.edu.cn}

\renewcommand{\shortauthors}{Fang et al.}

\begin{abstract}
   Improving the resolution of magnetic resonance (MR) image data is critical to computer-aided diagnosis and brain function analysis. Higher resolution helps to capture more detailed content, but typically induces to lower signal-to-noise ratio and longer scanning time. To this end, MR image super-resolution has become a widely-interested topic in recent times. Existing works establish extensive deep models with the conventional architectures based on convolutional neural networks (CNN). In this work, to further advance this research field, we make an early effort  to build a Transformer-based MR image super-resolution framework, with careful designs on exploring valuable domain prior knowledge. Specifically, we consider two-fold domain priors including the high-frequency structure prior and the inter-modality context prior, and establish a novel Transformer architecture, called Cross-modality high-frequency Transformer (Cohf-T), to introduce such priors into super-resolving the low-resolution (LR) MR images. 
  Experiments on two datasets indicate that Cohf-T achieves new state-of-the-art performance.
\end{abstract}



\maketitle

\begin{figure}[t]
\centering
\includegraphics[clip,width=1\linewidth]{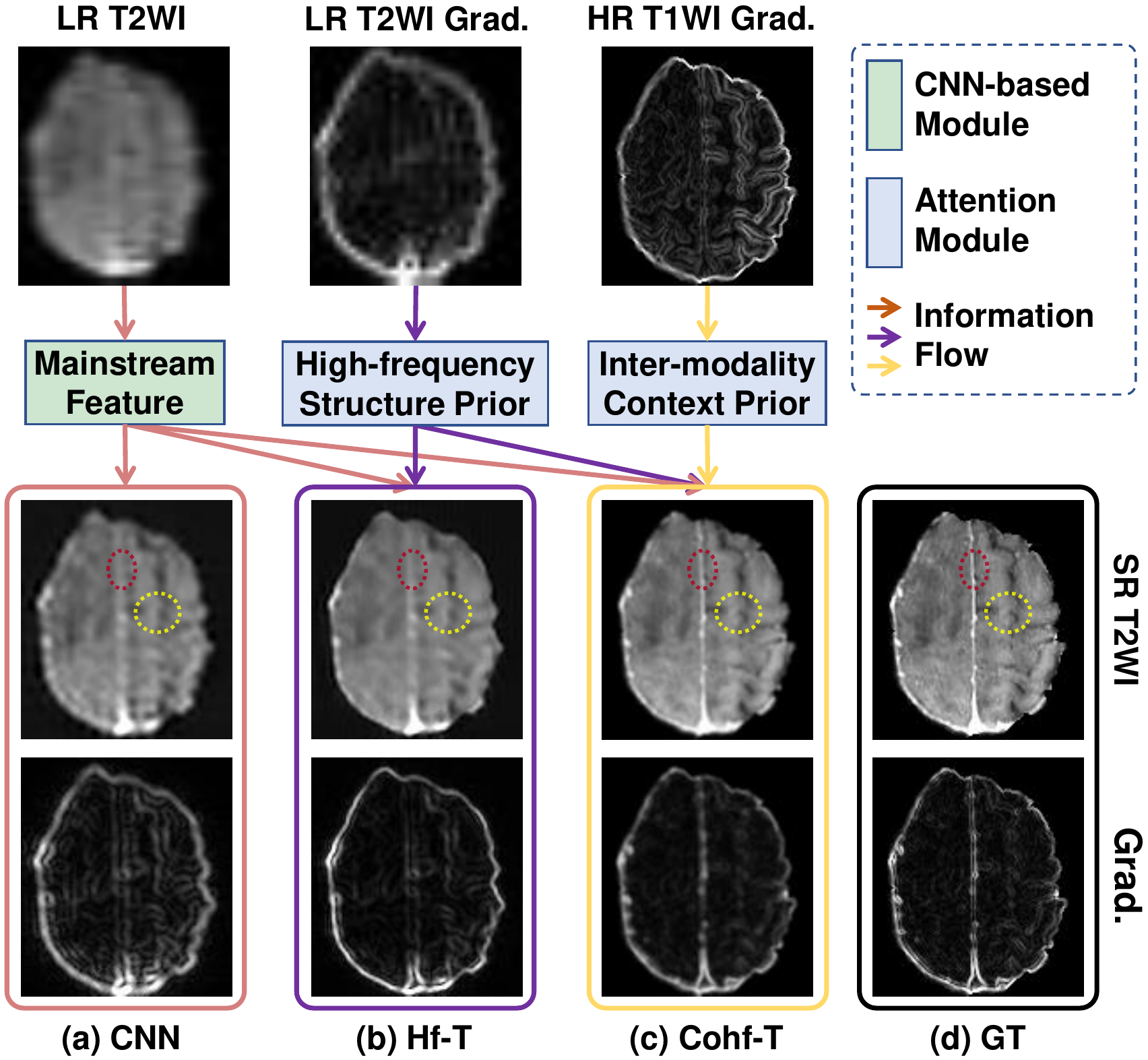}
\vspace{-6mm}
\caption{Aiming to super-resolve the input LR T2WI, we devise a Transformer-based framework capable of extracting both high-frequency structure prior and inter-modality context.
The mainstream of our method (a) is built upon convolutional neural networks. The high-frequency structure prior and inter-modality context improves the super-resolution performance, as indicated by Hf-T (High-frequency Transformer, (b))  and Cohf-T (Cross-modality high-frequency Transformer, (c)), respectively. The ground-truth (GT) HR T2WI and its gradient map are shown in (d).
}
\label{fig:teaser}
\vspace{-5mm}
\end{figure}

\section{Introduction}
\label{sec:intro}
Due to the superior capacity in capturing histopathological detail of soft tissues, magnetic resonance (MR) image becomes one of the most widely-used data in computer-aided diagnosis and brain function analysis. 
However, because of hardware and post-processing constraints, collecting MR images with higher resolution leads to lower signal-to-noise ratio or extends the scanning time~\cite{plenge2012super}.
For addressing this problem, one cost-effective way is to apply the super-resolution technology, which synthesizes the desired high-resolution (HR) MR image from the LR MR image. 

The current main-stream technique for MR image super-resolution (SR) is based on convolutional neural networks (CNN)~\cite{pham2017brain,zhao2019channel}. Although encouraging results are obtained by these CNN-based approaches, the intrinsic short-range reception mechanism is disadvantageous to the exploration of the global structure and long-range context information. This makes existing CNN-based MR image SR algorithms still sub-optimal for producing satisfactory results.

This paper makes an early effort to build a Transformer-based MR image super-resolution framework. In contrast to conventional CNN models, the Transformer model constituted by self-attention blocks has advantages in modeling the long-distance dependency \cite{vaswani2017attention}. Such a network architecture is firstly evolved in natural language processing (NLP) and then achieves enormous success in vision tasks like image recognition~\cite{dosovitskiy2020image} and semantic segmentation~\cite{strudel2021segmenter}. It is also verified that Transformer-based modelling is effective in medical image segmentation \cite{chen2021transunet} and registration \cite{lee2019image}. Here, we leverage it to advance the research on the super-resolution of T2-weighted MR image (T2WI), a typical MR data used in clinics but costing a very long scanning time.
 \begin{figure*}[h]
\centering
\includegraphics[clip,trim=2.0 2.0 1.2 2.0,width=0.95\linewidth]{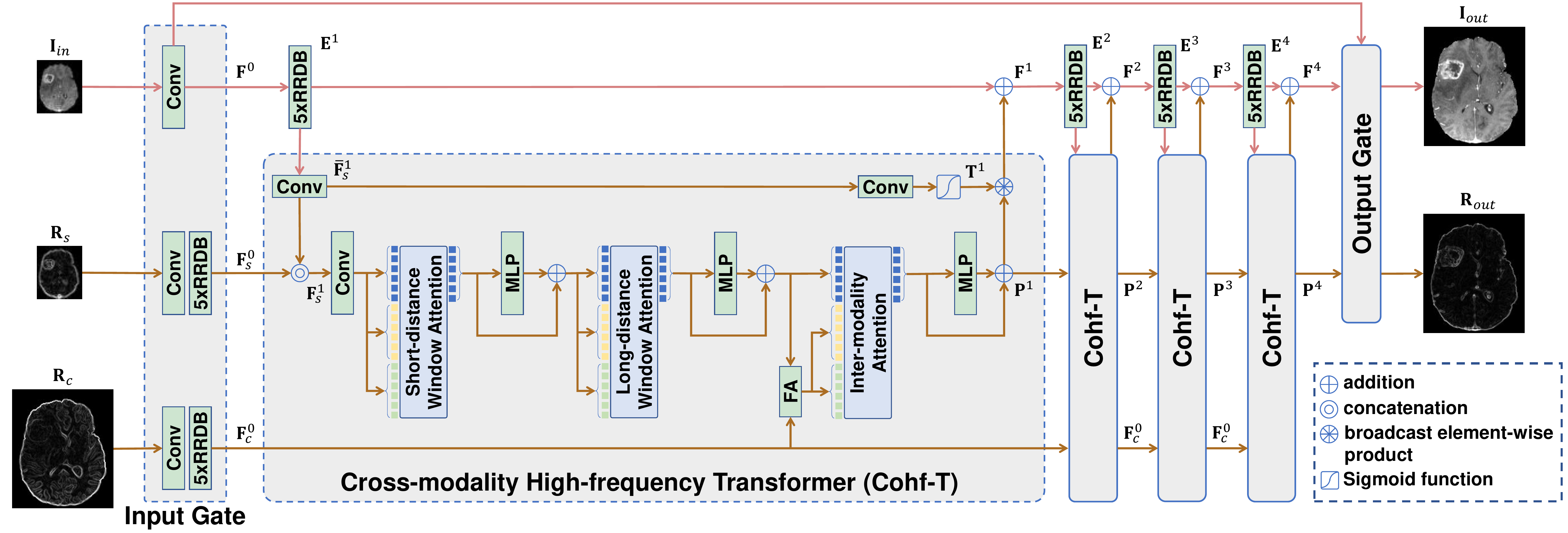}
\vspace{-3mm}
\caption{The pipeline of our proposed method. It consists of three main branches, a fully convolutional network for density domain super-resolution, a Transformer-based branch for restoring high-frequency signals in the gradient domain, and a guidance branch for extracting priors from the T1 modality. `Conv' and `RRDB' represent one $3\times3$ convolution layer and residual-in-residual dense block, respectively. `MLP' stands for multi-layer perceptron. 
}
\vspace{-4mm}
\label{fig:pipeline}
\end{figure*}
 
Existing transformer designs directly model self-attention in the original image domain. However, when implementing super-resolution on MR images, we find that the structure information held in the high-frequency domain would play a paramount role for model design as the organs appearing in the MR images usually share similar anatomical structures across persons, and the relation between different parts of the organ is regular~\cite{cherukuri2019deep} (see Fig. \ref{fig:teaser}). We term this as the high-frequency structure prior, and an example is provided to demonstrate the efficacy of the high-frequency structure prior in Fig.~\ref{fig:teaser} (b). To this end, the transformer model designed in this work performs on the high-frequency image gradients rather than the conventionally used original image pixels.

Another important domain knowledge is that when processing LR T2WI data, the high-resolution T1-weighted images (T1WI) can be used to provide rich inter-modality context priors, as 1) the complementary morphological information~\cite{zhang2020exploring,feng2021multi,zhang2021cross} captured by the T1WI can help infer the structural content of the T2WI , and 2) the acquisition of HR T1WI costs much less scanning time \footnote{\url{https://case.edu/med/neurology/NR/MRI\%20Basics.htm}}. An example is also provided to demonstrate the efficacy of such inter-modality context prior in Fig.~\ref{fig:teaser} (c).

To explore the domain priors mentioned above, we propose a novel Transformer architecture called Cross-modality high-frequency Transformer (Cohf-T). A novel learning framework is set up for super-resolving LR T2WI under the guidance of gradient maps of both LR T2WI and HR T1WI. As shown in Fig.~\ref{fig:pipeline}, our model has a main SR stream and a domain prior embedding stream, together with an input gate and an output gate. The domain prior embedding stream explores the high-frequency structure priors from the gradient map of LR T2WI and the inter-modality context from HR T1WI. Practically, both short-distance and long-distance dependencies are leveraged to explore high-frequency structure priors with the help of window attention modules.
Considering there exists distribution shift between T2WI and T1WI, an adaptive instance normalization module is devised for aligning their features before performing the cross-modality attention. Additionally, a novel basic attention module that encloses both intra-head and inter-head correlations is proposed to improve the relation extraction capacity of our Transformer-based framework. 

In summary, this work has three main contributions:
\begin{itemize}
\item We make an early effort to establish a Transformer-based framework for super-resolving T2-weighted MR images, based on the proposed Cross-modality high-frequency Transformer (Cohf-T).
\item In Cohf-T, we introduce the high-frequency structure prior and inter-modality context prior by designing novel intra-modality window attention and inter-modality attention modules.
\item Comprehensive experiments on two MR image super-resolution benchmarks demonstrate that the proposed Cohf-T achieves new state-of-the-art performance.
\end{itemize}

\section{Related Work}
\label{sec:related}
\subsection{MR Image Super-Resolution}
Improving the resolution of MR images is a long-lasting classical task in medical image analysis. Inspired by the rapid development in image super-resolution~\cite{dong2015image,ledig2017photo,zhang2018residual,wang2018esrgan,fang2019self,niu2020single,fang2022incremental} , CNN models have been the mainstream solutions  for the MR image super-resolution~\cite{chen2018brain,pham2017brain,zhao2019channel}. 
According to the current studies \cite{ma2020structure,yang2020learning,lu2021masa}, the low-frequency 
signals in MR data are relatively easy to reconstruct. In contrast, super-resolving the high-frequency signals, such as structures and textures, remains the main challenge.




Different from natural images, we can acquire MR images with multiple modalities via different imaging settings. A series of traditional algorithms~\cite{rousseau2010non,jain2017patch} attempt to explore prior context information from T1-weighted MR images for super-resolving T2-weighted or spectroscopy MR images.
\cite{iqbal2019super,feng2021multi,yang2021fast} further devise CNN models to exploit such inter-modality context information.
However, directly fusing features of multiple modalities via convolutions with small kernels can not sufficiently leverage the inter-modality dependencies. To further advance this research field, we devise a novel Transformer-based super-resolution framework. The proposed framework can capture long-distance dependencies for involving the high-frequency structure prior and inter-modality context prior, which is beyond the exploration of the existing works.  

\subsection{Transformer for Vision Tasks}
Transformer is primordially proposed for extracting long-distance relation context in NLP tasks~\cite{vaswani2017attention}. Recently, this technique has been extensively applied to computer vision tasks, considering it can help to make up the artifact of convolution that only local features can be captured with a limited kernel size \cite{chen2021pre,liang2021swinir,dosovitskiy2020image,strudel2021segmenter,peng2021conformer,liu2021swin}. 

Among the existing vision transformer models, self-attention~\cite{buades2011non,liu2018non,zhu2020pnen} and its variations, e.g., \cite{dai2019second,mei2021image}, are widely adopted to build the basic transformer block. For example, \cite{dai2019second} exploits the second-order attention based on covariance normalization for feature enhancement in image super-resolution. In~\cite{mei2021image}, the computation burden of the non-local operation is reduced by removing less informative correlations and constructing a sparse attention map. 
Transformer modules are also applied for tackling the multi-modal vision understanding tasks. Targeted at the multispectral object detection, \cite{qingyun2021cross} concatenates tokens from RGB and thermal modalities and then fuse them with symmetric attention modules.
\cite{liang2021cmtr} assigns specific modality
embeddings to tokens from different modalities for solving the visible-infrared person re-identification task.
Unlike these attention mechanisms, we introduce inter-head correlation in our attention modeling to further improve the feature interaction. Practically, three kinds of attention modules, namely short-distance window attention, long-distance window attention, and inter-modality attention, are incorporated in our transformer block.



\section{Method}
\label{sec:method}

\subsection{The Overall Learning Framework}
This paper is targeted at super-resolving low-resolution (LR) T2-weighted image (T2WI) under the guidance of high-resolution (HR) T1-weighted image (T1WI). Specifically, given the main input image, namely a LR T2WI $\mathbf I_{in} \in \mathbb R^{h \times w}$, we embed the high-frequency structure prior by calculating the gradient field for $\mathbf I_{in}$ and define the gradient image as the structure reference input (denoted by $\mathbf R_{s} = \sqrt{(\nabla_x \mathbf I_{in})^2+(\nabla_y \mathbf I_{in})^2+\epsilon}$, where $\epsilon$ is a constant and is set to $10^{-6}$). The gradient of the HR T1WI (denoted by $\mathbf R_{c}$) is regarded as an additional reference input for supplying inter-modality context information. 
 $h$ and $w$ denote the height and width of $\mathbf I_{in}$, respectively.
Then, a network architecture is built upon the cross-modality high-frequency transformer to predict the high-resolution T2-weighted image $\mathbf I_{out} \in \mathbb R^{rh \times rw}$. $r$ denotes the upsampling ratio. 

As shown in Fig.~\ref{fig:pipeline}, our proposed framework is composed of two streams, including the main super-resolution stream (performing on the LR T2WI $\mathbf I_{in}$) and the domain prior embedding stream, together with an input gate and an output gate.
The intensity levels of T1WI are not directly related to those of T2WI, e.g., the inflammation appears to be dark in T1WI but bright in T2WI. On the other hand, the two kinds of images share similar structures. 
Thus, we extract the domain prior information with the LR T2WI gradient image $\mathbf R_{s}$ and the HR T1WI gradient $\mathbf R_{c}$.


\vspace{1mm}
\noindent \textbf{Input Gate.} The input gate projects the input image data, including $\mathbf I_{in}$, $\mathbf R_{s}$, and $\mathbf R_{c}$ to their corresponding primary features $\mathbf F^0$, $\mathbf F_{s}^0$, and $\mathbf F_{c}^0$ through several convolutional layers and residual-in-residual dense blocks (RRDBs)~\cite{wang2018esrgan}.

\vspace{1mm}
\noindent \textbf{Main Super-Resolution Stream.}
There are four stages in the main super-resolution stream. In each stage, five RRDBs are utilized for deriving more complicated latent features. For the $i$-th stage, the extracted feature map is denoted by, $\mathbf E^i=\mathcal R(\mathbf F^{i-1})$. Then, $\mathbf E^i$ is fed into a cross-modality high-frequency transformer (Cofh-T) block to interact with the high-frequency structure prior and inter-modality context prior, resulting in a domain prior embedding $\mathbf P^i=\text{Cohf-T}(\mathbf E^i,\mathbf F_s^i,\mathbf F_c^0)$. A more detailed description of the Cohf-T block can be referred to in the next subsection.
Extra structural information is extracted from $\mathbf E^i$ via a convolution layer, resulting in $\bar{\mathbf F}_s^i =\mathcal C(\mathbf E^i)$. Then, a series of attention modules are employed to explore domain priors (namely $\mathbf P^i$) from $\mathbf F_s^i$, $\bar{\mathbf F}_s^i$, and $\mathbf F_c^0$.
Finally, the prior-induced latent features of the current stage are produced via a shortcut connection: $\mathbf F^i=\mathbf E^i+\mathbf T^i \circ \mathbf P^i$, where $\mathbf T^i$ is a single-channel selection map inferred from $\bar{\mathbf F}_s^i$, and `$\circ$' is the broadcast element-wise production.

\vspace{1mm}
\noindent \textbf{Domain Prior Embedding Stream.} This network stream which is composed of four cascaded Cohf-T blocks involves the $\mathbf F_s^0$ and $\mathbf F_c^0$ as inputs. In particular, for the $i$-th Cohf-T block, we first combine previous prior features and additional structure features acquired from main stream features as $\mathbf F_s^{i}=\mathcal C(\mathbf P^{i-1}||\bar{\mathbf F}_s^i)$\footnote{For the first Cohf-T block, $\mathbf F_s^{1}=\mathcal C(\mathbf P^{0}||\bar{\mathbf F}_s^1)$ where $\mathbf P^{0}=\mathbf F_s^0$.}. 
Then, short-distance window attention and long-distance window attention are sequentially integrated to extract the high-frequency structure knowledge from $\mathbf F_s^{i}$. Next, together with inter-modality context prior features $\mathbf F_c^{0}$, the obtained attention features are further passed through an inter-modality attention module to generate the final domain prior embedding features $\mathbf P^{i}$.

\begin{figure*}[t]
\centering
\includegraphics[width=1\linewidth]{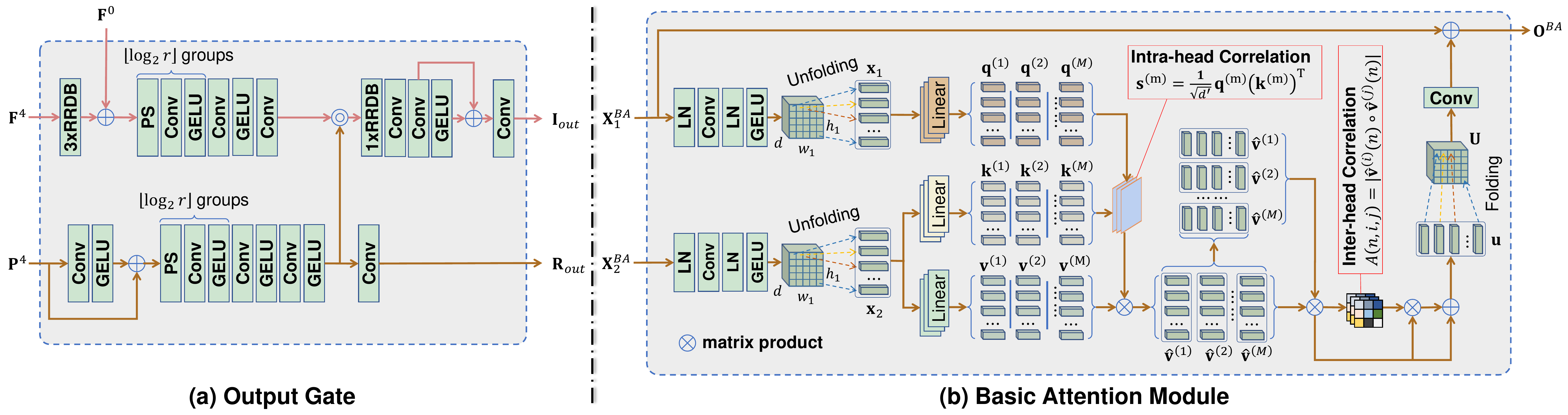}
\vspace{-6mm}
\caption{Details of the output gate (a) and basic attention module (b). In (a), `PS' denotes the pixel shuffle operation, and `GELU' denotes the Gaussian error linear unit~\cite{hendrycks2016gaussian}. In (b),  `LN' stands for layer normalization; `Unfolding' aggregates features of local patches into patch-level representations; `Folding' is the reverse operation of `unfolding'.
}
\vspace{-3mm}
\label{fig:exit}
\end{figure*}

\vspace{1mm}
\noindent \textbf{Output Gate.}
Based on the latent feature $\mathbf F^4$ extracted from the main super-resolution stream and the domain prior embedding feature ${\mathbf P}^4$, an output gate is built up for jointly synthesizing high-resolution intensity and gradient images. The concrete architecture is shown in Fig.~\ref{fig:exit} (a). 

\vspace{1mm}
\noindent \textbf{Objective Function.}
We adopt the mean square error and structural similarity index measure to estimate the consistency between network predictions and the ground-truths. The following objective function is adopted for constraining the main prediction $\mathbf I_{out}$,
\begin{equation} \label{eq:loss-density}
    L_{in} = \alpha \textrm{MSE}(\mathbf I_{out}, \mathbf I_{gt}) - (1-\alpha) (\textrm{SSIM}(\mathbf I_{out}, \mathbf I_{gt})),
\end{equation}
where $\alpha$ is a weighting coefficient, and $\mathbf I_{gt}$ represents the ground-truth HR T2WI. $\textrm{MSE}(\cdot)$ denotes the mean square error function, and $\textrm{SSIM}(\cdot)$ denotes the function for calculating the structural similarity index measure.

The similar objective function is adopted for calculating the training loss for the gradient prediction $\mathbf R_{out}$,
\begin{equation} \label{eq:loss-grad}
    L_c = \alpha \textrm{MSE}(\mathbf R_{out}, \mathbf R_{gt}) - (1-\alpha) \textrm{SSIM}(\mathbf R_{out}, \mathbf R_{gt}),
\end{equation}
where $\mathbf R_{gt}= \sqrt{(\nabla_x \mathbf I_{gt})^2+(\nabla_y \mathbf I_{gt})^2+\epsilon}$. The overall objective function is formed by combining (\ref{eq:loss-density}) and (\ref{eq:loss-grad}), $L=L_{in}+\lambda L_c$. $\lambda$ is a weighting coefficient for the gradient restoration constraint. 

\subsection{Main Designs in Cofh-T}

\subsubsection{The Basic Attention Module} \label{sec:attention}
Different from the classic multi-head self-attention (MHSA) that is widely used in the existing transformer models~\cite{vaswani2017attention,dosovitskiy2020image}, we propose an alternative attention module as the basic unit of our model by taking both intra-head and inter-head correlations into consideration.

Given an input feature map $\mathbf X_1^{BA} \in \mathbb R^{ h_1 \times w_1 \times d}$ and a reference context feature map $\mathbf X_2^{BA} \in \mathbb R^{h_2 \times w_2 \times d }$ ($\mathbf X_2^{BA}$ might be equal to $\mathbf X_1^{BA}$), the target of the attention module is to extract the well-aligned context information from $\mathbf X_2^{BA}$ for enhancing the representation of $\mathbf X_1^{BA}$. The whole calculation process is illustrated in Fig.~\ref{fig:exit} (b).

First, the layer normalization is applied to separately standardize  $\mathbf X_1^{BA}$ and $\mathbf X_2^{BA}$. 
A $1\times1$ convolution accompanied with the other layer normalization and activation function GELU is utilized to embed $\mathbf X_1^{BA}$ into a $d$-dimensional feature map. Afterwards, it is decomposed into local patches with size of $p\times p$ and then arranged into patch-wise representation $\mathbf x_1^{BA}\in \mathbb R^{N\times dp^2}$ ($N=h_1w_1/p^2$) via the unfolding operation which flattens local patches in the feature map into vectors.
The similar process is utilized to transform $\mathbf X_2^{BA}$ to patch-wise representation $\mathbf x_2^{BA}$, in which the size and stride of the convolution are both set to $\rho \times \rho$ for registering the spatial dimensions of $\mathbf X_2^{BA}$ with those of $\mathbf X_1^{BA}$, i.e., $\rho=h_2/h_1=w_2/w_1$.
$M$ groups of linear layers are used for generating $M$ query representations denoted by $\mathop \cup_{m=1}^M \mathbf q^{(m)} \in\mathbb R^{N \times d^\prime}$ ($d^\prime=dp^2/M$) from $\mathbf x_1^{BA}$. 
Key representations (denoted by $\mathop \cup_{m=1}^M \mathbf k^{(m)} \in\mathbb R^{N\times d^\prime}$) and value representations (denoted by $\mathop \cup_{m=1}^M \mathbf v^{(m)} \in \mathbb R^{N\times d^\prime}$) are calculated by feeding $\mathbf x_2$ into another $2M$ linear layers. 

Then, like the classic MHSA, $M$ intra-head correlation maps are calculated by the softmax function. The value at $(i,j)$ of the $m$-th correlation map $\textbf{s}^{(m)}$ is estimated as, 
\begin{equation}
    s^{(m)}_{i,j}=\frac{\exp(\|q^{(m)}_i \circ k^{(m)}_j\|/\sqrt{d^\prime})} {\sum_{j^\prime=1}^N \exp(\|q^{(m)}_i \circ k^{(m)}_{j^\prime}\|/\sqrt{d^\prime})},
\end{equation}
where $q^{(m)}_i$ and $k^{(m)}_j$ represent the $i$-th and the $j$-th rows of $\textbf{q}^{(m)}$ and $\textbf{k}^{(m)}$, respectively, and $\|\cdot\|$ denotes the summation of all elements in the input tensor.
These correlation maps estimate point-wise dependencies between two input feature maps from the same heads. With the correlation encoded by $\mathbf s^{(m)}$, we renew the value feature by, $\hat{\mathbf v}^{(m)}=\mathbf s^{(m)}\mathbf v^{(m)}$.

To explore the dependencies among different heads, we devise an inter-head correlation modelling algorithm. First, the inter-head correlation matrix $\mathbf A\in\mathbb R^{N\times M\times M}$ is estimated as below,
\begin{equation}
     A_{n,i,j}=\frac{\exp(\|\hat{v}^{(i)}_n \circ \hat{v}^{(j)}_n\|)}
    { \sum_{j=1}^m \exp(\|\hat{v}^{(i)}_n \circ \hat{v}^{(j)}_n\|)},
\end{equation}
where $\hat{v}^{(i)}_n$ represents the $n$-th row of $\hat{\textbf{v}}^{(i)}$.
Then, the value features are combined with the inter-head correlation, resulting in $\mathop \cup_{m=1}^M \mathbf u^{(m)} \in \mathbb R^{N\times d^\prime}$. The $n$-th row of $\mathbf u^{(m)}$ is calculated as,
\begin{equation}
u^{(m)}_n = \sum_{j=1}^M (1+ A_{n,m,j}) \hat{v}^{(j)}_n,
\end{equation}
where $\mathbb I\in \mathbb R^{N\times M\times M}$ is formed by stacking $N$ unit matrices with size of $M\times M$. 
Afterwards, $\mathbf u^{(m)}$-s are transformed into a $h_1\times w_1 \times d$ tensor denoted by $\mathbf U$ with the folding operation\footnote{The folding operation first expands the first dimension of $\mathbf u$ into the spatial size of $h_1/p\times w_1/p$ and then allocates the $p^2d$-dimensional feature vector of every point into a $p\times p \times d$ patch.}. 
Finally, a $3\times3$ convolution layer is adopted to post-process $\mathbf U$ which is then fused into $\mathbf X_1^{BA}$ via a skip connection, deriving of an enhanced variant of $\mathbf X_1^{BA}$ (namely $\mathbf O^{BA}$). 
\subsubsection{Intra-modality Window Attention}
Short-distance and long-distance windows are adopted to model the intra-modality dependency for restoring the high-resolution gradient field. In particular, the short-distance structure helps to amend local details such as noncontinuous boundaries with surrounding structure information. At the same time, the long-distance structure can leverage the repetition of textural and structural patterns to enhance the pixel-wise feature representation. 

\begin{figure}
\centering
\includegraphics[width=1\linewidth]{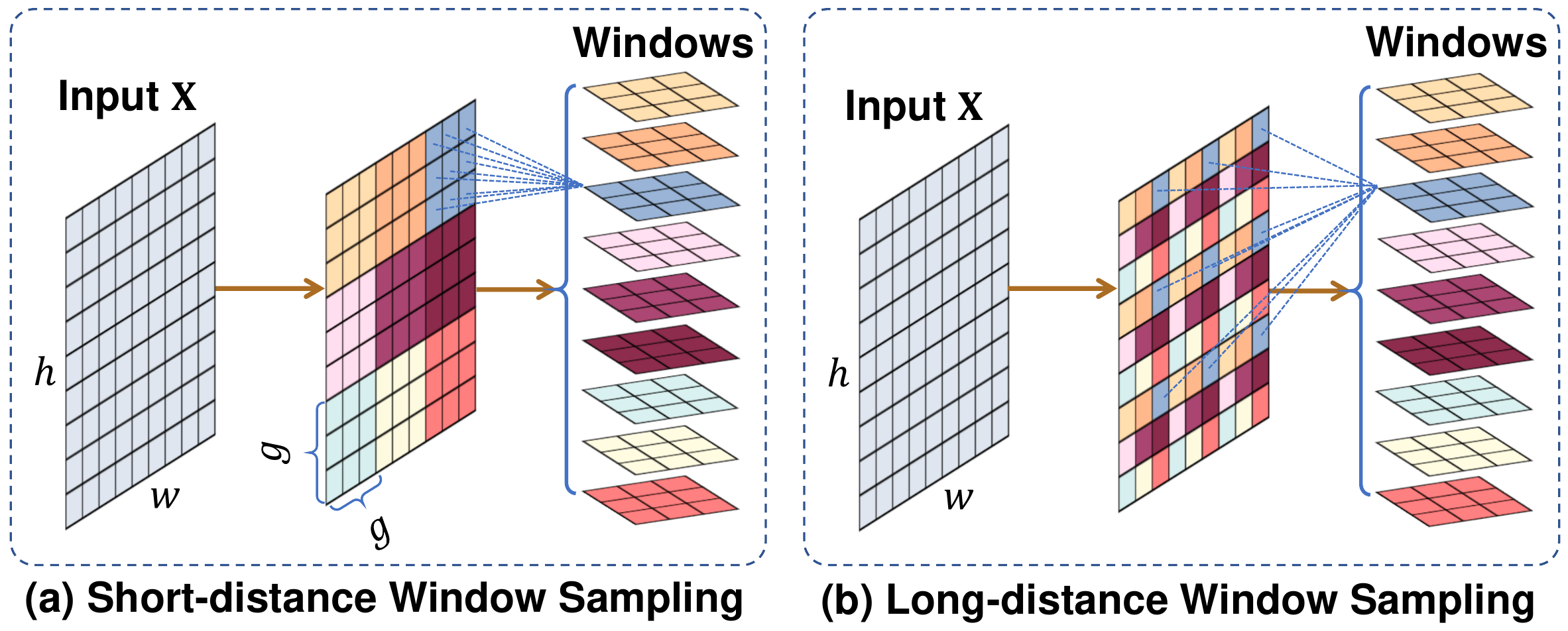}
\vspace{-6mm}
\caption{Two window sampling designs are employed for exploring short-distance (a) and long-distance (b) dependencies, respectively.
}
\vspace{-3mm}
\label{fig:window}
\end{figure}

Inspired from~\cite{liu2021swin}, for acquiring the short-distance structure, we uniformly decompose the input feature map $\mathbf X \in \mathbb R^{h\times w\times d}$ into compact windows with the size of $g\times g$ as in Fig.~\ref{fig:window} (a).
Regarding the input feature map itself as the reference context features, the attention module in Section~\ref{sec:attention} is applied for enhancing features of every window. 
Afterward, the features are processed with a residual multi-layer perceptron (MLP) with a structure of `LN-Conv1x1-LN-GELU-Conv1$\times$1' (see Cohf-T in Fig.~\ref{fig:pipeline}). 

 
For extracting the long-distance dependency information, we propose to sample dilated windows from the input feature map (as shown in Fig.~\ref{fig:window} (b)). The horizontal or vertical distance between neighboring points in each window is $w/g$ or $h/g$.
Then, the attention module, together with a residual MLP, is employed to enhance the features of each long-distance window. Once $g$ is larger than $w/g$ and $h/g$, cascading the short-distance and long-distance window attention modules can efficiently extract context information from the full image. The former module aggregates the context information from the surrounding $g\times g$ neighborhood, while the latter module involves the knowledge of all windows. 



\subsubsection{Inter-modality Attention}
\begin{figure}
\centering
\includegraphics[width=1\linewidth]{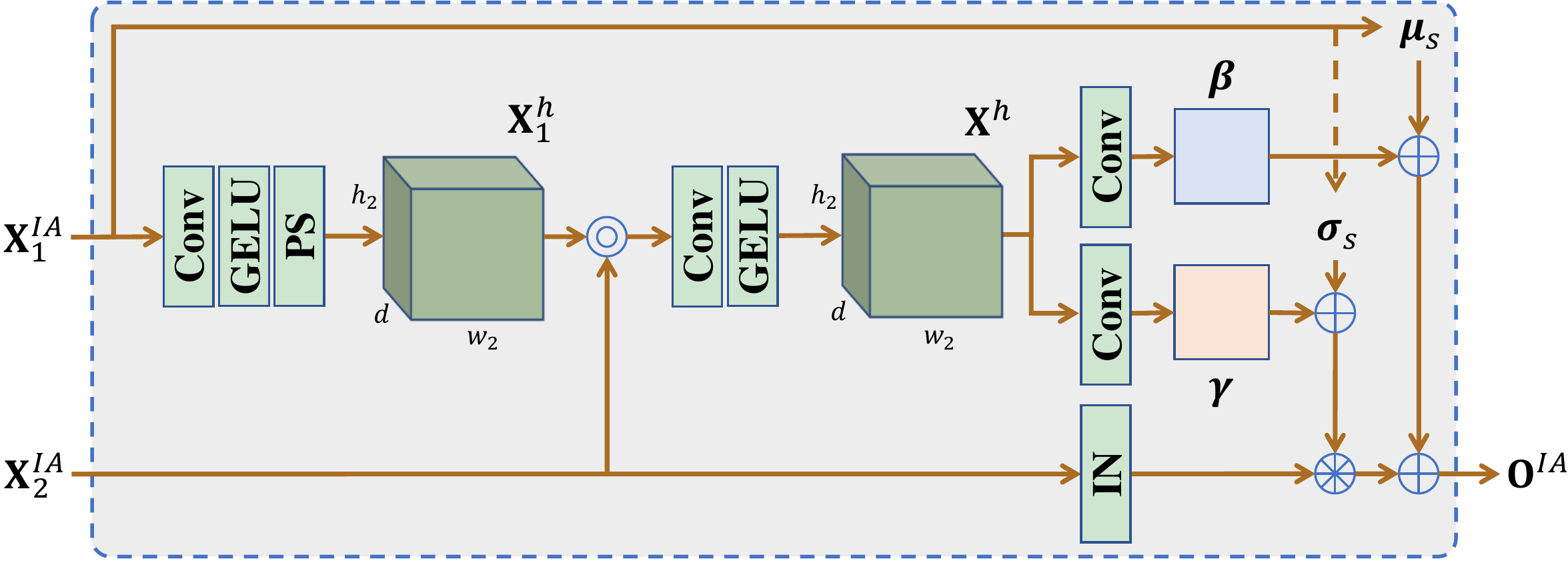}
\vspace{-6mm}
\caption{We devise a cross-modality feature adaptation module for aligning the feature distributions between two modalities.
}
\vspace{-3mm}
\label{fig:adapt}
\end{figure}

To involve the valuable inter-modality context information, we devise a cross-modality attention module within each Cohf-T block. Considering that there would be a large domain gap between the gradient fields of different modality data, directly injecting features of the T1 modality into those of the T2 modality would not be the optimal solution. Inspired from~\cite{huang2017arbitrary,lu2021masa}, we adopt a point-wise adaptive instance normalization scheme to reduce the cross-modality representation gap as shown in Fig. \ref{fig:adapt}. 

Given the input feature maps $\mathbf X_1^{IA} \in \mathbb R^{h\times w\times d}$ and $\mathbf X_2^{IA} \in \mathbb R^{rh\times rw\times d}$, the goal is to transfer the content of $\mathbf X_2^{IA}$ to a feature space that is finely aligned to $\mathbf X_1^{IA}$. We estimate channel-wise mean $\boldsymbol \mu_2\in\mathbb R^d$  and variance $\boldsymbol \sigma_2 \in \mathbb R^d$ from $\mathbf X_2^{IA}$, and then apply the instance normalization to standardize $\mathbf X_2^{IA}$, resulting in $\mathbf X_2^\prime$ as below,
\begin{equation}
\mathbf X_2^\prime [x,y,j] =\frac{\mathbf X_2^{IA}[x,y,j]-\boldsymbol \mu_2[j]}{ \boldsymbol \sigma_2[j]}.
\end{equation}
To estimate $\boldsymbol{\beta}$ and $\boldsymbol{\gamma}$, we first align the height and width of $\mathbf X_1^{IA}$ with those of $\mathbf X_2^{IA}$. Specifically, one convolution layer is first used to increase the channel number of $\mathbf X_1^{IA}$ to $dr^2$. Then, the resulted feature map is arranged to obtain $\mathbf X_1^{h} \in \mathbb R^{rh\times rw\times d}$ via the pixel shuffle operation. Afterward, $\mathbf X_2^{IA}$ and  $\mathbf X_1^{h}$ are concatenated and compressed into a $rh\times rw\times d$ tensor (denoted by $\mathbf X^{h}$) via one convolution layer. Point-wise affine parameters $\boldsymbol \beta \in \mathbb R^{rh\times rw \times 1}$ and $\boldsymbol \gamma \in \mathbb R^{rh\times rw \times 1}$ are generated from $\mathbf X^{h}$ with separate convolution layers. We can also calculate the channel-wise mean ($\boldsymbol \mu_1$) and variance ($\boldsymbol \sigma_1$) vectors of $\mathbf X_1^{IA}$. Then, the module outputs $\mathbf O^{IA}$ as the updating of $\mathbf X_2^{IA}$:
\begin{equation}
    \mathbf O^{IA}[x,y,j] = \mathbf X_2^\prime[x,y,j] (\boldsymbol \sigma_1[j]+\boldsymbol \gamma[x,y])+\boldsymbol \mu_1[j]+\boldsymbol \beta[x,y].
    \label{adapt}
\end{equation}
The adaptive normalization in~\cite{lu2021masa} calculates global affine parameters are calculated in each batch normalization layer. Our devised normalization to differs from it in the adoption of point-wise affine parameters, which model local distribution deviation more finely.

The adaptive normalization helps project features of T1 modality to get close to the feature distribution of the T2 modality. The feature streams of $\mathbf R_c$ and $\mathbf R_s$ correspond to $\mathbf X_2^{IA}$ and $\mathbf X_1^{IA}$, respectively.
Subsequently, the cross-modality attention is performed by using $\mathbf X_1^{IA}$ to generate the query features and using $\mathbf O^{IA}$ to create the key and value features. Then, another residual MLP is attached for further feature enhancement.
\begin{table*}[t]
\centering
\fontsize{9}{10}\selectfont
\setlength{\tabcolsep}{1.8mm}{
    \begin{tabular}{l| c|c| c|c |c|c |c|c |c|c |c|c }
        \toprule
        \multirow{2}{*}{Method} &
        \multicolumn{6}{c|}{BraTS2018 dataset} & \multicolumn{6}{c}{IXI dataset} \\ \cmidrule(l){2-7} \cmidrule(l){8-13}
        
        & \multicolumn{2}{c|}{2$\times$} & \multicolumn{2}{c|}{3$\times$} & \multicolumn{2}{c|}{4$\times$} &
        \multicolumn{2}{c|}{2$\times$} & \multicolumn{2}{c|}{3$\times$} & \multicolumn{2}{c}{4$\times$} 
        \\ 
        \cmidrule(l){2-3} \cmidrule(l){4-5} \cmidrule(l){6-7} \cmidrule(l){8-9} \cmidrule(l){10-11} \cmidrule(l){12-13}
                 & PSNR           & SSIM              & PSNR             & SSIM                   & PSNR           & SSIM & PSNR           & SSIM & PSNR           & SSIM & PSNR           & SSIM \\ \midrule
        Bicubic  & 30.92 & 0.9198 & 28.19   & 0.9062 & 25.87 & 0.8388 & 29.71 & 0.9067 & 26.01   & 0.8070 & 24.32 & 0.7296  \\ 
        RDN~\cite{zhang2018residual}                       & 34.57 & 0.9536 & 32.70   & 0.9324 & 30.92 & 0.9172 & 33.54 & 0.9435 & 31.02   & 0.9307 & 29.89 & 0.8998  \\
        RCAN~\cite{zhang2018image}                       & 34.82 & 0.9556 & 33.11   & 0.9310 & 30.88 & 0.9301 & 33.66 & 0.9422 & 30.79   & 0.9302  & 29.60 & 0.8936 \\
        SAN~\cite{dai2019second}                       & 34.81 & 0.9544 & 32.91   & 0.9332 & 30.90 & 0.9281 & 34.31 & 0.9513 & 31.28   & 0.9401 & 29.59 & 0.8825 \\
        CFSR~\cite{tian2020coarse} & 34.79 & 0.9566 & 32.66   & 0.9299 & 30.77 & 0.9290 & 34.24 & 0.9521 & 31.30   & 0.9320 & 29.78 & 0.8874  \\
        HAN~\cite{niu2020single}                       & 35.01 & 0.9572 & 33.19   & 0.9351 & 31.03 & 0.9262 & 34.10 & 0.9480 & 31.36   & 0.9377  & 29.96 & 0.8993  \\
        SRResCGAN~\cite{umer2020deep} & 34.21 & 0.9492 & 33.00   & 0.9388  & 30.99 & 0.9048 & 34.02 & 0.9522 & 31.21 & 0.9388 & 30.24 & 0.8924  \\
        SPSR~\cite{ma2020structure}                       & 35.11 & 0.9585 & 33.32   & 0.9370 & 31.11 & 0.9264 & 34.47 & 0.9511 & 31.61   & 0.9382  & 30.47 & 0.8931 \\
        SMSR~\cite{wang2021exploring} & 35.31 & 0.9601 & 33.62   & 0.9402 & 31.32 & 0.9342 & 34.77 & 0.9588 & 31.88   & 0.9477 & 30.36 & 0.9004  \\
        NLSN~\cite{mei2021image} & 35.43 & 0.9655 & 33.99   & 0.9423 & 31.72 & 0.9321 & 34.51 & 0.9577 & 31.65   & 0.9432 & 30.64 & 0.8859 \\ 
        SwinIR~\cite{liang2021swinir} & 35.42 & 0.9519 & 33.87   & 0.9425  & 31.84 & 0.9386 & 34.62 & 0.9401 & 31.83   & 0.9385 & 30.91 & 0.8915 \\ 
        TTSR~\cite{yang2020learning} & - & - & - & - & 31.33 & 0.9322 & - & - & - & - & 30.63 & 0.9044  \\
        MASA-SR~\cite{lu2021masa} & - & - & - & - & 31.44 & 0.9357 & - & - & - & - & 30.78 & 0.9032 \\ 
        MINet~\cite{feng2021multi} & 35.11 & 0.9587 & 33.25   & 0.9377 & 31.24 & 0.9355 & 34.34 & 0.9529 & 31.33 & 0.9208 & 30.58 & 0.8987 \\
        T2-Net~\cite{feng2021task}                       & 35.32 & 0.9590 & 33.68   & 0.9420 & 31.64 & 0.9377 & 34.46 & 0.9532 & 31.60   & 0.9421 & 30.40 & 0.8992 \\
        MTrans~\cite{feng2021mtrans} & 35.35 & 0.9595 & 33.73   & 0.9411 & 31.73 & 0.9320 & 34.67 & 0.9556 & 31.69   & 0.9410 & 30.44 & 0.8993 \\
         \midrule
        Ours-S & 36.10 & 0.9681 & 34.29   & 0.9460 & 32.25 & 0.9392  & 35.17 & 0.9610 & 32.29   & 0.9490 & 31.45 & 0.9093\\ 
        Ours-M & 36.29 & 0.9698 & 34.43   & 0.9488 & 32.66 & 0.9401   & 35.41 & 0.9627 & 32.51   & 0.9491 & 31.48 & 0.9125 \\
        Ours-L & \textbf{36.85} & \textbf{0.9739} & \textbf{34.84}   & \textbf{0.9507} & \textbf{33.26} & \textbf{0.9425}  & \textbf{35.73} & \textbf{0.9645} & \textbf{32.79}   & \textbf{0.9498} & \textbf{31.82} & \textbf{0.9129}\\
        \bottomrule
    \end{tabular}
}
\caption{Comparison with existing methods on BraTS2018 and IXI datasets, under $2\times$, $3\times$, and $4\times$ upsampling settings.}
\label{tab:brats}
\vspace{-3mm}
\end{table*}

\section{Experiments}
\label{sec:exper}
\subsection{Datasets and Evaluation Metrics}

\begin{figure*}
\centering
\includegraphics[clip,trim=2.0 2.0 1.2 2.0,width=0.98\linewidth]{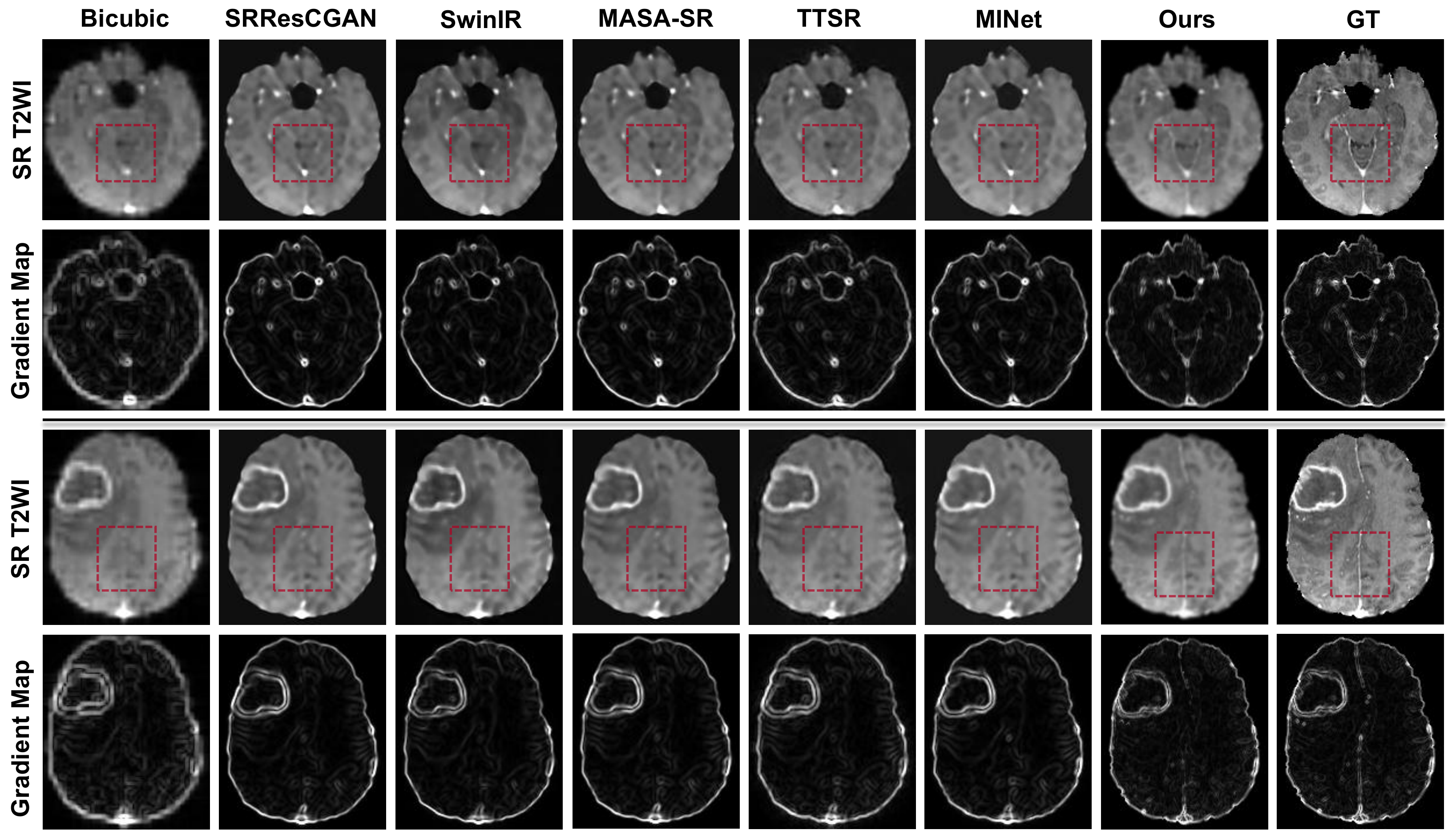}
\vspace{-3mm}
\caption{Qualitative comparison with other methods on $4\times$ image super-resolution.}
\vspace{-3mm}
\label{fig:cmpothers}
\end{figure*}
\begin{figure}
\centering
\includegraphics[width=0.95\linewidth]{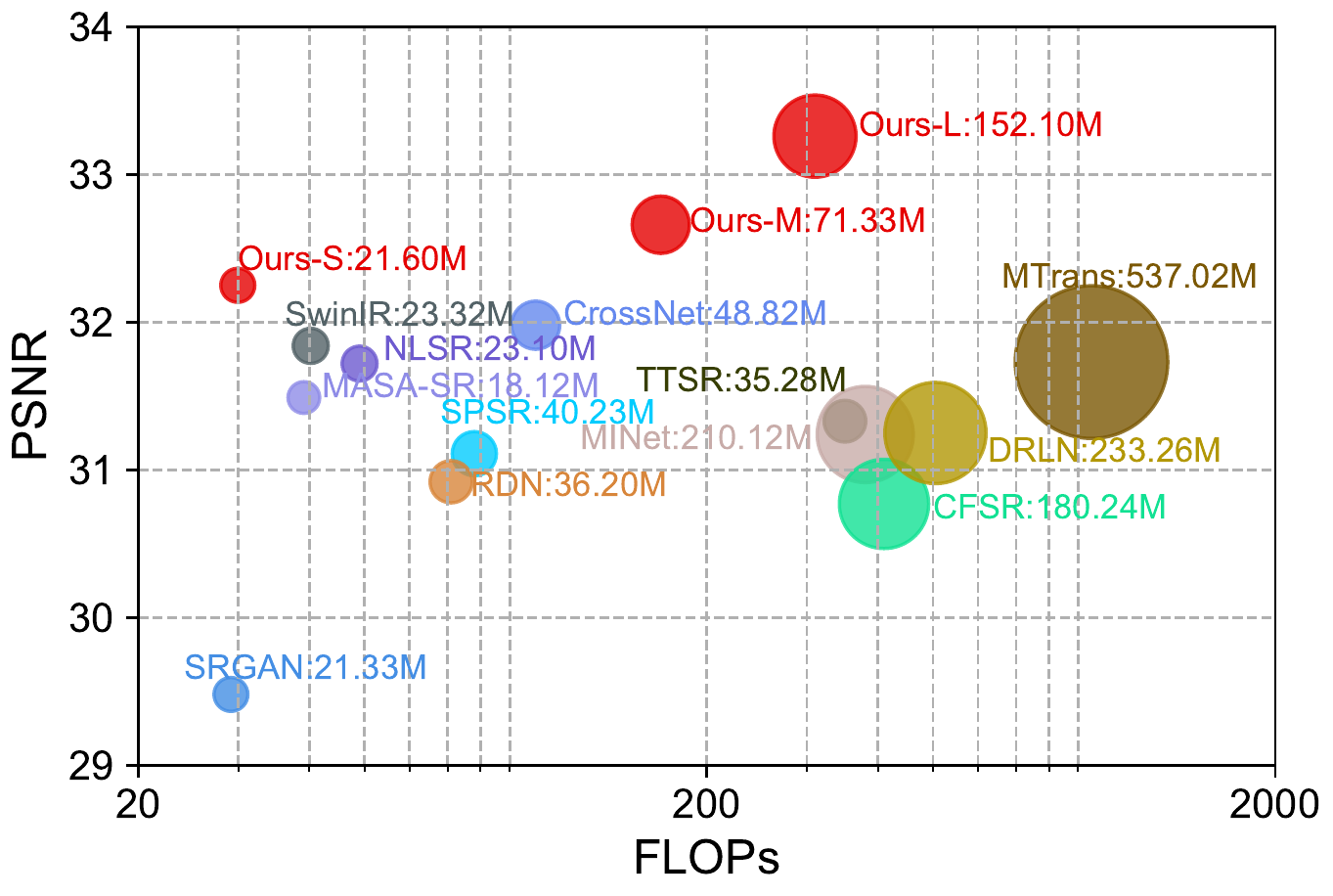}
\vspace{-5mm}
\caption{Visualisation of PSNR, FLOPS, and model size of image super-resolution algorithms. All metrics are evaluated in the $4\times$ super-resolution task on BraTS2018 dataset. 
}
\vspace{-3mm}
\label{fig:flops}
\end{figure}

Two multi-modal MR image datasets are employed for validating super-resolution algorithms. 
\begin{itemize}
\item \textbf{BraTS2018} is composed of 750 MR volumes~\cite{menze2014multimodal}. They are registered to a uniform anatomical template and interpolated into the same resolution (1mm$\times$1mm$\times$1mm). They are split into 484 volumes (including 75,020 images) for training, 66 volumes (including 10,230 images) for validating, and 200 volumes (including 31,000 images) for testing. The width and height of all images are both 240.
\item \textbf{IXI} is composed of 576 MR volumes collected from three hospitals in London, including Hammersmith Hospital using a Philips 3T system, Guy’s Hospital using a Philips 1.5T system, and Institute of Psychiatry using a GE 1.5T system. They are split into 404 volumes (including 48,480 images) for training, 42 volumes (including 5,040 images) for validating, and 130 volumes (including 15,600 images) for testing. The width and height of all images are both 256.
\end{itemize}

We follow \cite{feng2021mtrans} to synthesize low-resolution T2WIs. PSNR and SSIM are used for performance evaluation.
\vspace{-1mm}
\subsection{Implementation Details}
\vspace{-1mm}
Our method is implemented under PyTorch with a 32GB V100 GPU. Adam is used for network optimization, and the weight decay is set to $10^{-4}$. 
The network is trained for 400 epochs with a mini-batch size of 4.
The learning rate is initially set to $10^{-4}$ and decayed by half every 100 epochs.
Without specification, the feature dimension $d$ is set to 32; in the intra-modality window attention module, the input feature map is decomposed into  windows with the size of $6\times 6$ (namely $g=6$), and $p$ is set to 1;
in the inter-modality attention module, $p$ is set to 5; 
$M$ is set to 4; other parameters are set as: $\lambda=0.5$ and $\alpha=0.95$. 


\subsection{Comparison with Other Methods}
We compare our method against extensive existing super-resolution methods including~\cite{feng2021task,feng2021mtrans,feng2021multi,yang2020learning,lu2021masa,zhang2018residual,zhang2018image,dai2019second,tian2020coarse,niu2020single,umer2020deep,ma2020structure,wang2021exploring,mei2021image,liang2021swinir}.
For~\cite{yang2020learning,lu2021masa}, the T1WI is regarded as the reference image. We implement three variants of our method with different model sizes, including: 1) For 'Ours-S', $d$ is set to 16, two feature extraction and enhancement stages are used, and each RRDB block only contains two RDBs composed of three convolutions; 2) For 'Our-M',  $d$ is set to 16, three feature extraction and enhancement stages are used, and each RRDB block only contains three RDBs composed of three convolutions; 3) `Our-L' is the final variant of our method with default settings. 
Experiments on BraTS2018 and IXI datasets are presented in Table~\ref{tab:brats}.

\vspace{1mm}
\noindent \textbf{Quantitative Comparisons.}
On BraTS2018 dataset, our small variant `Our-S' performs better than the best natural image super-resolution method SwinIR~\cite{liang2021swinir} by 0.41dB, and the best reference-based image super-resolution method MASA-SR~\cite{lu2021masa} by 0.81dB 
under the $4\times$ unsampling setting, while consuming fewer parameters and less memory cost (see Fig.~\ref{fig:flops}). Compared to existing state-of-the-art MR image super-resolution method MTrans~\cite{feng2021mtrans}, our final variant `Ours-L' gives rise to PSNR gains of 4.2\%, 3.3\%, and 4.8\% under $2\times$, $3\times$, and $4\times$ unsampling settings respectively.
On IXI dataset, our method performs better than other methods as well. Particularly, it derives results with 3.1\%, 3.5\%, and 4.5\% higher PSNR than the results of MTrans, under $2\times$, $3\times$, and $4\times$ upsampling settings, respectively.

\vspace{1mm}
\noindent \textbf{Qualitative Comparisons} against other methods including SRResGAN~\cite{umer2020deep}, SwinIR~\cite{liang2021swinir}, TTSR~\cite{yang2020learning}, MASA-SR and MINet are presented in Fig.~\ref{fig:cmpothers}. 
Compared to images super-resolved by other methods, the results of our approach have more accurate local details. 
The gradient maps indicate that our method can generate HR T2WI with sharper and more complete structures.

\vspace{1mm}

\noindent \textbf{Model Complexity.}
The model sizes and memory consumption of different methods are illustrated in Fig.~\ref{fig:flops}. Overall, our method outperforms other super-resolution methods with fewer parameters and less memory cost. 

\begin{figure*}[h]
\centering
\includegraphics[clip,trim=4.0 4.0 4.0 4.0,width=0.24\linewidth]{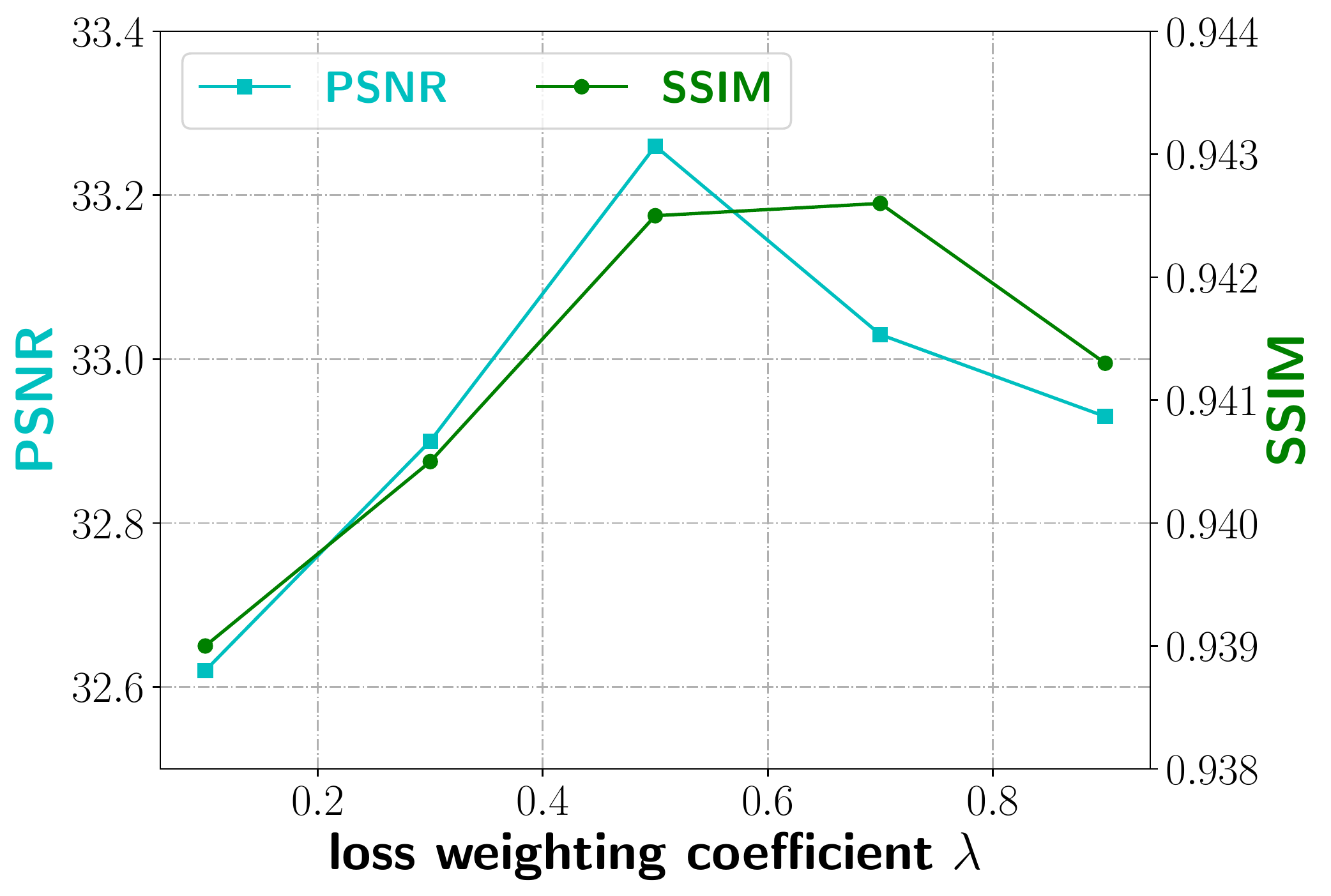}
\includegraphics[width=0.24\linewidth]{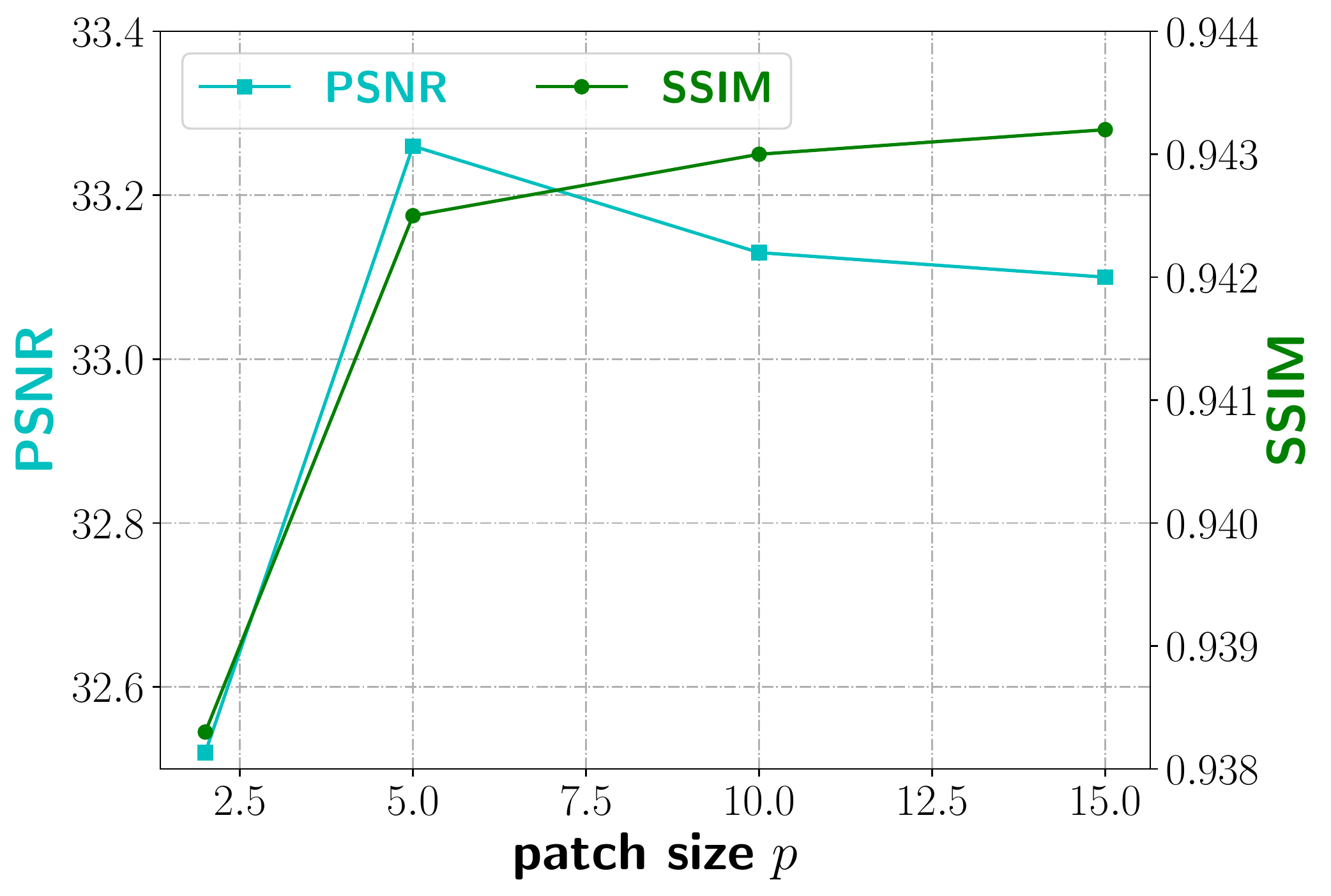}
\includegraphics[width=0.24\linewidth]{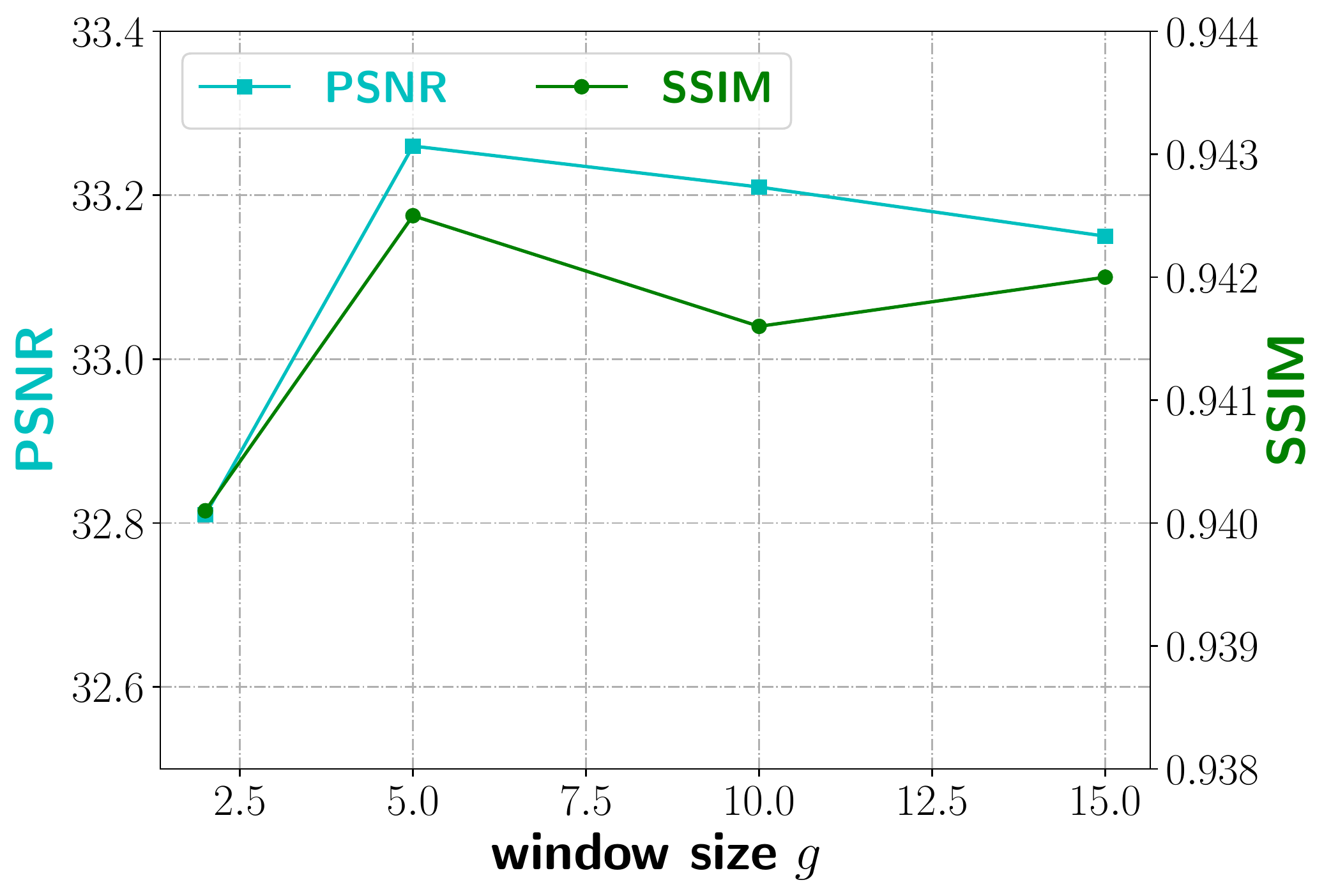}
\includegraphics[width=0.24\linewidth]{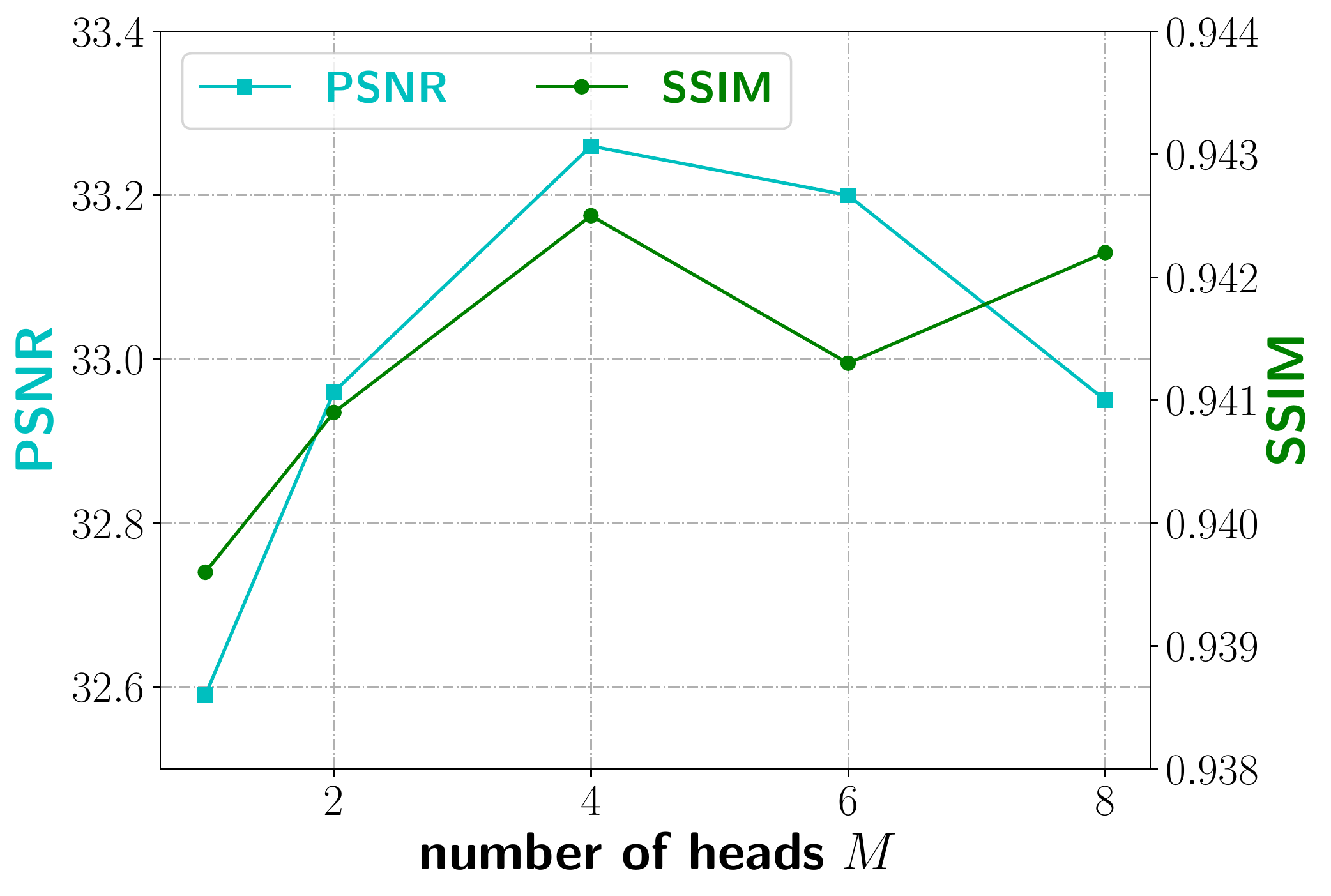}
\vspace{-3mm}
\caption{Performance of using different parameters. From left to right: the loss weighting factor $\lambda$; patch size $p$ in the inter-modality attention module; window size $g$ in intra-modality window attention modules; the number of attention heads $M$. }
\label{fig:para}
\end{figure*}
\subsection{Ablation Study}
Extensive ablation studies are conducted on the BraTS2018 dataset under the $4\times$ upsampling setting to validate the efficacy of critical components in our method.

\begin{table}[t]
\centering
\fontsize{9}{10}\selectfont
\setlength{\tabcolsep}{3mm}{
    \begin{tabular}{l| c|c}
        \toprule
        Variant & PSNR & SSIM \\ \midrule
        baseline  & 31.06 & 0.9225 \\
        w/o S-IntraM-WA or L-IntraM-WA & 32.20 & 0.9306 \\
        w/o S-IntraM-WA & 32.46 &  0.9362\\
        w/o L-IntraM-WA & 32.54 & 0.9380\\
        w/o InterM-A & 32.17 & 0.9345\\
        w/o InterH-Corr & 32.69 & 0.9406\\
        w/o AdaptIN & 32.77 & 0.9398 \\
        full version & 33.26 & 0.9425\\
        \bottomrule
    \end{tabular} 
}
\caption{Ablation study on attention modules. `S/L-IntraM-WA': short-distance/long-distance intra-modality window attention; `InterM-A': inter-modality attention; `InterH-Corr': inter-head correlation; `AdaptIN': adaptive instance normalization.}
\label{tab:ab-att}
\vspace{-5mm}
\end{table}
\begin{figure}[t]
\includegraphics[clip,trim=0 0 0 0,width=1.0\linewidth]{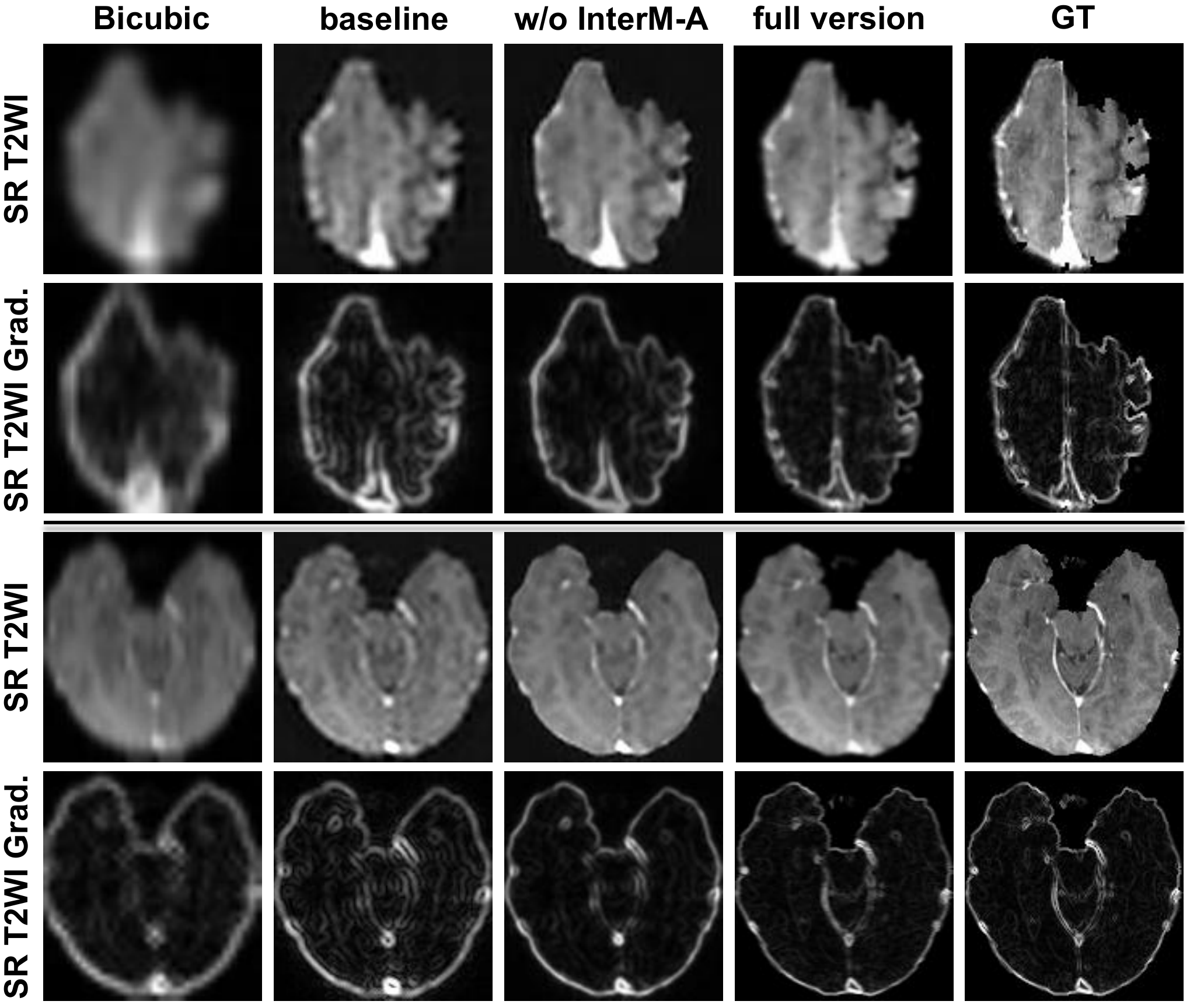}
\vspace{-6mm}
\captionof{figure}{Qualitative comparison between different variants of our method. The baseline is the CNN-based mainstream branch. 
`w/o InterM-A' only utilizes the  gradient map of T2WI for extracting structure priors.  }\label{fig:cmpinner}
\vspace{-2mm}
\end{figure}

\vspace{1mm}
\noindent \textbf{Efficacy of Attention Modules.} We implement variants of our method by removing the attention modules or their inner units. The experimental results are reported in Table~\ref{tab:ab-att}. The baseline model is formed by removing all intra-modality window attention and inter-modality attention modules. Without using the inter-modality attention (encoded as `w/o InterM-A'), the PSNR is dramatically decreased by 1.09dB compared to the full version of our method. 
Qualitative comparisons are provided in Fig.~\ref{fig:cmpinner} to illustrate the efficacy of using inter-modality context information. As can be observed, after incorporating the gradient map of the T1WI with our devised inter-modality attention module, the structural information is recovered more completely. 
When both short-distance and long-distance intra-modality window attentions are abandoned (encoded as `w/o S-IntraM-WA or L-IntraM-WA'), the PSNR is decreased to 32.20dB, which is 1.06dB lower than that of the full version. Removing any separate short-distance (w/o S-IntraM-W) or long-distance (w/o L-IntraM-WA) window attention leads to performance degradation, which indicates that short-distance and long-distance dependencies have complementary effects in extracting the high-frequency structure priors. 
Without using the adaptive instance normalization (w/o AdaptIN) for aligning feature distributions across modalities, the PSNR is decreased by 0.49dB.
The removing of the inter-head correlations (w/o InterH-Cor) in the basic attention module induces to 0.57dB PSNR reduction. 
\begin{table}[t]
\centering
\fontsize{9}{10}\selectfont
\setlength{\tabcolsep}{3mm}{
    \begin{tabular}{l|l|c|c|c}
        \toprule
        Domain    & Arch.  &\#Params & PSNR &  SSIM \\ \midrule
        Input & CNN-RB &      171.127M   & 31.86 & 0.9377 \\
        Input & Cohf-T & 152.106M & 33.10 & 0.9409 \\
        Gradient & CNN-RB  & 171.127M & 31.90 & 0.9383 \\
        Gradient & Cohf-T & 152.106M & 33.26 & 0.9425 \\
        \bottomrule
    \end{tabular}
}
\caption{Performance of applying different designs for domain prior embedding. `CNN-RB' means  the CNN-based residual block is used to replace the Cohf-T. }
\label{tab:ab-prior}
\vspace{-5mm}
\end{table}

\vspace{1mm}
\noindent \textbf{Different Designs for Domain Prior Embedding.} We apply different designs for exploring high-frequency structure priors and inter-modality context priors as in Table~\ref{tab:ab-prior}. 
We try to replace the Cofh-T with CNN-based residual blocks to capture intra-modality and inter-modality dependencies.
Though more parameters are used, CNN-based residual blocks perform worse than Cofh-T (e.g., having 1.36dB lower PSNR under the gradient domain). It is also validated that the original input domain is sub-optimal to the exploration of domain prior information. Performing domain prior embedding in the original input domain produces results with 0.16dB lower PSNR than the gradient domain when Cofh-T is used for attention modeling.

\begin{table}[t]
\centering
\fontsize{9}{10}\selectfont
\setlength{\tabcolsep}{4mm}{
    \begin{tabular}{l|c|c|c}
        \toprule
        Metric  & w/o $L_c$ & w/o SSIM Loss & Final Loss \\ \midrule
        PSNR    & 32.33  & 32.86   & 33.26 \\
        SSIM    & 0.9410 & 0.9377  & 0.9425 \\
        \bottomrule
    \end{tabular}
}
\caption{Efficacy of loss components. }
\label{tab:ab-loss}
\vspace{-3mm}
\end{table}

\vspace{1mm}
\noindent \textbf{Efficacy of Loss Components.} According to the results in Table~\ref{tab:ab-loss}, the gradient level loss can improve the PSNR value while preserving the SSIM value. The SSIM loss is adopted for measuring the dissimilarity between local structures of predicted and ground-truth images. It can bring marginal gains to the SSIM value. 

\vspace{1mm}
\noindent \textbf{Choice of Hyper-parameters.}
The impact of hyper-parameters on the performance of our method is demonstrated in Fig.~\ref{fig:para}, from which we can observe that: 1) Our method achieves the best performance when the loss weighting coefficient $\lambda$ is set to around 0.5; 2) For the inter-modality attention module, using too small {patch size $p$} would be unfavorable to constructing reliable cross-modality dependencies, while the performance tends to be saturated after $p$ reaches 5; 
3) For intra-modality window attention modules, larger {window size $g$} would benefit to a more thorough exploration of relational structure priors. However, the performance saturates when $g$ becomes larger than 5; 4) Adopting more attention heads helps extract diversified types of relations. The performance increase as the \textbf{$M$} grows within 4 and tends to be saturated when $M>4$.



\section{Conclusion}
\label{sec:conc}
\vspace{-1mm}
In this paper, we devise a novel Transformer-based framework to tackle the MR image super-resolution task. 
A Cross-modality high-frequency Transformer (Cohf-T) module is proposed for exploring structure priors from the gradient domain and inter-modality context from an additional modality. 
We extract intra-modality and inter-modality dependencies to capture the domain priors via the Transformer module.
The inter-head correlations can bring extra promotion to feature enhancement in attention modules. The distribution alignment strategy based on adaptive instance normalization is beneficial for fusing features of different modalities.
Experiments on two datasets demonstrate that our method achieves state-of-the-art performance on the MR image super-resolution task.

\section*{ACKNOWLEDGMENTS}
\vspace{-1mm}
This work was supported in part by  the National Natural Science Foundation
of China (No. 62003256, 62106235, 61876140, 62027813, U1801265, and U21B2048), in part by Open Research Projects of Zhejiang Lab (No.
2019kD0AD01/010), and in part by the Exploratory Research Project under Grant 2022PG0AN01.


\end{document}